\pdfoutput=1

\documentclass[11pt]{article}

\usepackage[]{acl}

\usepackage{times}
\usepackage{latexsym}
\usepackage{amsmath}

\usepackage[T1]{fontenc}

\usepackage[utf8]{inputenc}

\usepackage{microtype}

\usepackage{todonotes}

\title{The Lifecycle of "Facts": A Survey of Social Bias in Knowledge Graphs}

\author{Angelie Kraft\textsuperscript{{\normalfont 1}} \and Ricardo Usbeck\textsuperscript{{\normalfont 12}} \\
    \textsuperscript{1}Department of Informatics, Universität Hamburg, Germany\\
    \textsuperscript{2}Hamburger Informatik Technologie-Center e.V. (HITeC), Germany \\
    {\{angelie.kraft, ricardo.usbeck\}@uni-hamburg.de}}

\begin{document}
\maketitle

\begin{abstract}
Knowledge graphs are increasingly used in a plethora of downstream tasks or in the augmentation of statistical models to improve factuality. However, social biases are engraved in these representations and propagate downstream. We conducted a critical analysis of literature concerning biases at different steps of a knowledge graph lifecycle. We investigated factors introducing bias, as well as the biases that are rendered by knowledge graphs and their embedded versions afterward. Limitations of existing measurement and mitigation strategies are discussed and paths forward are proposed.
\end{abstract}

\section{Introduction}

Knowledge graphs (KGs) provide a structured and transparent form of information representation and lie at the core of popular Semantic Web technologies. They are utilized as a source of truth in a variety of downstream tasks (e.g., information extraction~\cite{DBLP:journals/semweb/Martinez-Rodriguez20}, link prediction~\citep{getoor2007introduction,DBLP:journals/ki/NgomoSGHDLOS21}, or question-answering~\cite{DBLP:journals/semweb/HoffnerWMULN17,DBLP:journals/kais/DiefenbachLSM18,DBLP:journals/widm/ChakrabortyLMTL21,DBLP:conf/sigir/JiangU22}) and in hybrid AI systems (e.g., knowledge-augmented language models~\cite{DBLP:conf/emnlp/PetersNLSJSS19,DBLP:conf/coling/SunSQGHHZ20,yu2022survey} or conversational AI~\cite{DBLP:conf/sigir/GaoG018,DBLP:conf/ictir/GerritseHV20}). In the latter, KGs are employed to enhance the factuality of statistical models
~\cite{DBLP:conf/www/AthreyaNU18,DBLP:journals/corr/abs-2204-09149}. In this overview article, we question the ethical integrity of these facts and investigate the lifecycle of KGs \citep{DBLP:conf/semweb/AuerBDEHILMMNSTW12,DBLP:journals/semweb/Paulheim17} with respect to bias influences.\footnote{We focus on the KG lifecycle from a bias and fairness lens. For reference, the processes investigated in Section \ref{sources_creation} correspond to the \textit{authoring stage} in the taxonomy by \citet{DBLP:conf/semweb/AuerBDEHILMMNSTW12}. The representation issues in KGs (Section \ref{bias_in_KG}) and KG embeddings (Sections \ref{bias-in-KGE} and \ref{bias-mitigation-in-KGE}) which affect downstream task bias relate to Auer et al.'s \textit{classification stage}.} 

\begin{figure}[htp!]
    \includegraphics[width=\columnwidth]{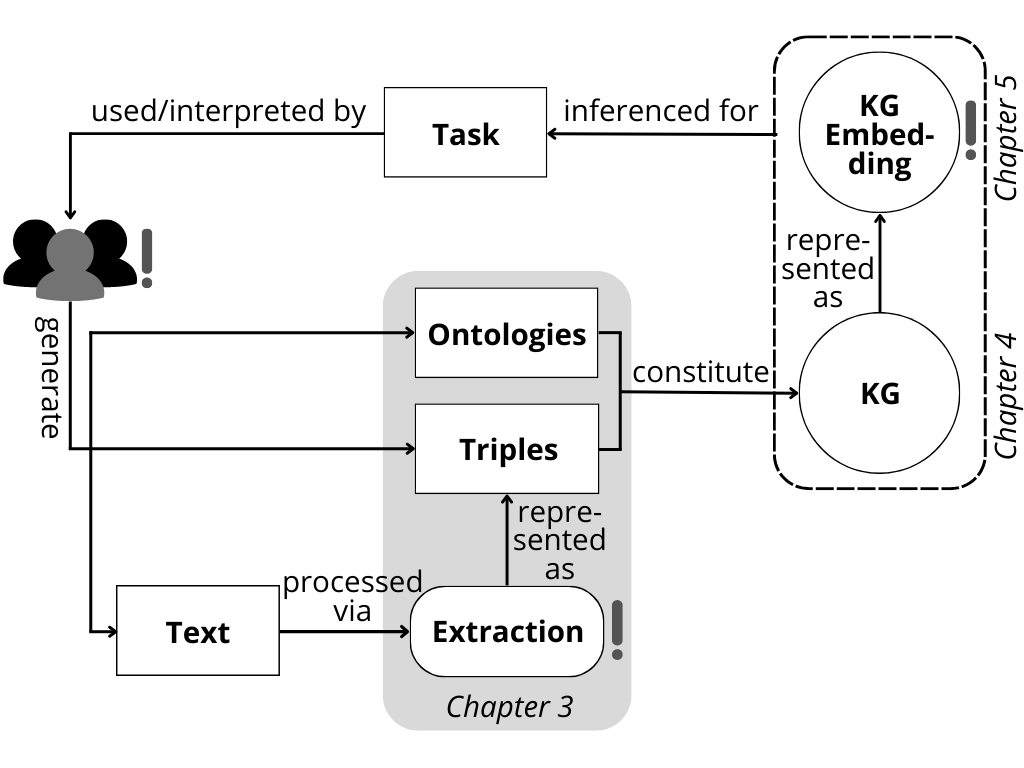}
    \caption{Overview of the knowledge graph lifecycle as discussed in this paper. Exclamation marks indicate factors that introduce or amplify bias. We examine bias-inducing factors of triple crowd-sourcing, hand-crafted ontologies, and automated information extraction (Chapter 3), as well as the resulting social biases in KGs (Chapter 4) and KG embeddings, including approaches for measurement and mitigation (Chapter 5).}
    \label{fig:schema_lifecycle}
\end{figure}

We claim that KGs manifest social biases and potentially propagate harmful prejudices. To utilize the full potential of KG technologies, such ethical risks must be targeted and avoided during development and application. Using an extensive literature analysis, this article provides a reflection on previous efforts and suggestions for future work.

We collected articles via Google Scholar\footnote{A literature search on Science Direct, ACM Digital Library, and Springer did not provide additional results.} and filtered for titles including \textit{knowledge graph/base/resource}, \textit{ontologies}, \textit{named entity recognition}, or \textit{relation extraction},  paired with variants of \textit{bias}, \textit{debiasing}, \textit{harms}, \textit{ethical}, and \textit{fairness}. We selected peer-reviewed publications (in journals, conference or workshop proceedings, and book chapters) from 2010 onward, related to social bias in the KG lifecycle. This resulted in a final count of 18 papers. Table \ref{emb-bias-overview} gives an overview of the reviewed works and
Figure \ref{fig:schema_lifecycle} illustrates the analyzed lifecycle stages. 

\section{Notes on Bias, Fairness, and Factuality}
\label{theory}
In the following, we clarify our operational definitions of the most relevant concepts in our analysis. 

\subsection{Bias}

If we refer to a model or representation as \textit{biased}, we --- unless otherwise specified --- mean that the model or representation is \textit{socially biased}, i.e., biased towards certain social groups. This  is usually indicated by a systematic and unfairly discriminating deviation in the way members of these groups are represented compared to others \citep{friedmann1996bias} (also known as \textit{algorithmic bias}).
Such bias can stem from pre-existing societal inequalities and attitudes, such as prejudice and stereotypes, or arise on an algorithmic level, through design choices and formalization \citep{friedmann1996bias}.  
From a more impact-focused perspective, algorithmic bias can be described as "a skew that [causes] harm" (Kate Crawford, Keynote at NIPS2017). Such harm can manifest itself in unfair distribution of resources or derogatory misrepresentation of a disfavored group. We refer to \textit{fairness} as the absence of bias.

\subsection{Unwanted Biases and Harms}
\label{harms}

One can distinguish between \textit{allocational} and \textit{representational harms} \citep[Barocas et al., as cited in,][]{blodgett2020language}, where the first refers to the unfair distribution of chances and resources and the second more broadly denotes types of insult or derogation, distorted representation, or lack of representation altogether.
To quantify biases that lead to representational harm, analyses of more abstract constructs are required.  \citet{mehrabi_bias_ml_2022}, for example,  measure indicators of representational harm via \textit{polarized perceptions}: a predominant association of groups with either negative or positive prejudice, denigration, or favoritism. Polarized perceptions are assumed to correspond to societal stereotypes. They can \textit{overgeneralize} to all members of a social group (e.g., "\textit{all} lawyers are dishonest"). 
It can be said that harm is to be prevented by avoiding or removing algorithmic bias. However, different views on the conditions for fairness can be found in the literature and, in consequence, different definitions of \textit{unwanted} bias. 

\subsection{Factuality versus Fairness}

We consider a KG factual if it is representative of the real world. For example, if it contains only male U.S. presidents, it truthfully represents the world as it is and has been. 
However, inference based on this snapshot would lead to the prediction that people of other genders cannot or will not become presidents. This would  be false with respect to U.S. law and/or undermine the potential of non-male persons. Statistical inference over historical entities is one of the main usages of KGs. The factuality narrative, thus, risks consolidating and propagating pre-existing societal inequalities and works against matters of social fairness. Even if the data represented are not affected by sampling errors, they are restricted to describing \textit{the world as it is} as opposed to \textit{the world as it should be}. We strive for the latter kind of inference basis. Apart from that, in the following sections we will learn that popular KGs are indeed affected by sampling biases, which further amplify societal biases.

\section{Entering the Lifecycle: Bias in Knowledge Graph Creation}
\label{sources_creation}
We enter the lifecycle view (Figure \ref{fig:schema_lifecycle}) by investigating the processes underlying the creation of KGs.  We focus on the human factors behind the authoring of \textit{ontologies} and \textit{triples} which constitute KGs. Furthermore, we address automated  \textit{information extraction}, i.e., the detection and extraction of entities and relations from text, since these approaches can be subject to algorithmic bias. 

\subsection{Triples: Crowd-Sourcing of Facts}

Popular large-scale KGs, like Wikidata~\cite{DBLP:journals/cacm/VrandecicK14} and DBpedia~\cite{DBLP:conf/semweb/AuerBKLCI07} are the products of continuous crowd-sourcing efforts. Both of these examples are closely related to Wikipedia, where the top five languages (English, Cebuano, German, Swedish, and French) constitute 35\% of all articles on this platform.\footnote{\url{https://en.wikipedia.org/wiki/List_of_Wikipedias}}  It can be said that Wikipedia is Euro-centric in tendency. Moreover, the majority of authors are white males.\footnote{\url{https://en.wikipedia.org/wiki/Gender_bias_on_Wikipedia;  https://en.wikipedia.org/wiki/Racial_bias_on_Wikipedia}} As a result, the data transport a particular homogeneous set of interests and knowledge \citep{beytia2021visual, DBLP:conf/icwsm/WagnerGJS15}. This \textit{sampling bias} affects the geospatial coverage of information \citep{janowicz_debiasing_2018} and leads to higher barriers for female personalities to receive a biographic entry \citep{beytia2021visual}. In an experiment, \citet{demartini_implicit_2019} asked crowd contributors to provide a factual answer to the (politically charged) question of whether or not Catalonia is a part of Spain. The diverging responses indicated that participants' beliefs of what counts as true differed largely. This is an example of bias that is beyond a subliminal psychological level. In this case, structural aspects like consumed media and social discourse play an important role. To counter this problem, \citet{demartini_implicit_2019} suggests actively asking contributors for evidence supporting their statements, as well as keeping track of their demographic backgrounds. This makes underlying motivations and possible sources for bias traceable.  

\subsection{Ontologies: Manual Creation of Rules}
Ontologies determine rules regarding allowed types of entities and relations or their usage. They are often hand-made and  a source of bias \citep{janowicz_debiasing_2018}
due to the influence of opinions, motivations, and personal choices \citep{keet2021exploration}: Factors like scientific opinions (e.g., historical ideas about race), socio-culture (e.g., how many people a person can be married to), or political and religious views (e.g., classifying a person of type X as a \textit{terrorist} or a \textit{protestor}) can proximately lead to an encoding of social bias. Also structural constraints like the ontologies' granularity levels can induce bias \citep{keet2021exploration}. Furthermore, issues can arise from the types of information used to characterize a person entity. Whether one attributes the person with their skin color or not could theoretically determine the emergence of racist bias in a downstream application \citep{paparidis-towards-2021}. 
\citet{geller-2021-detecting} give a practical example for detection and alleviation of ontology bias in a real-world scenario. The authors discovered that ontological gaps in the medical context lead to an under-reporting of race-specific incidents. They were able to suggest countermeasures based on a structured analysis of real incidents and external terminological resources.


\subsection{Extraction: Automated Extraction of Information}
\label{auto_info_extraction}
Natural language processing (NLP) methods can be used to recognize and extract entities (named entity recognition; NER) and their relations (relation extraction; RE), which are then represented as $[$head entity, relation, tail entity$]$ tuples (or as $[$subject, predicate, object$]$, respectively). 

\citet{mehrabi_man_2020} showed that the NER system CoreNLP  \citep{manning-etal-2014-stanford} exhibits binary gender bias. They used a number of template sentences, like "<Name> is going to school" or "<Name> is a person" using male and female names\footnote{While most of the works presented here refer to gender as a binary concept, this does not agree with our understanding. We acknowledge that gender is continuous and technology must do this reality justice.} from 139 years of census data. The model returned more erroneous tags for female names. Similarly, \citet{mishra_assessing_2020} created synthetic sentences from adjusted Winogender \citep{DBLP:conf/naacl/RudingerNLD18} templates with names associated with different ethnicities and genders. A range of different NER systems were evaluated (bidirectional LSTMs with Conditional Random Field (BiLSTM CRF) \citep{DBLP:journals/corr/HuangXY15} on GloVe \citep{DBLP:conf/emnlp/PenningtonSM14}, ConceptNet \citep{DBLP:conf/aaai/SpeerCH17} and ELMo \citep{DBLP:conf/acl/PetersABP17} embeddings, CoreNLP, and spaCy\footnote{\url{https://spacy.io/}} NER models). Across models, non-white names yielded on average lower performance scores than white names. Generally, ELMo exhibited the least bias. Although ConceptNet is debiased for gender and ethnicity\footnote{https://blog.conceptnet.io/posts/2017/conceptnet-numberbatch-17-04-better-less-stereotyped-word-vectors/}, it was found to produce strongly varied accuracy values.

\citet{gaut_towards_2020} analyzed binary gender bias in a popular open-source neural relation extraction (NRE) model, OpenNRE \citep{han-etal-2019-opennre}. For this purpose, the authors created a new dataset, named WikiGenderBias (sourced from Wikipedia and DBpedia). All sentences describe a gendered subject with one of four relations: \textit{spouse}, \textit{hypernym}, \textit{birthData}, or \textit{birthPlace} (DBpedia mostly uses occupation-related hypernyms).
The most notable bias found was the spouse relation. It was more reliably predicted for male than female entities. This observation stands in contrast to the predominance of female instances with spouse relation in WikiGenderBias. The authors experimented with three different mitigation strategies: downsampling the training data to equalize the number of male and female instances, augmenting the data by artificially introducing new female instances, and finally word embedding debiasing \citep{bolukbasi2016man}. Only downsampling facilitated a reduction of bias that did not come at the cost of model performance. 

Nowadays, contextualized transformer-based encoders are used in various NLP applications, including NER and NRE. Several works have analyzed the various societal biases encoded in large-scale word embeddings (like word2vec~\citep{word2vec2013mikolov,bolukbasi2016man} or BERT~\citep{devlin2018bert,kurita2019measuring}) or language models (like GPT-2~\citep{gpt2,DBLP:conf/nips/KirkJVIBDSA21} and GPT-3 \citep{gpt3,abid2021large}). Thus, it is likely that these biases also affect the downstream tasks discussed here. \citet{li2021robustness} used two types of tasks to analyze bias in BERT-based RE on the newly created Wiki80 and TACRED \citep{zhang-etal-2017-position} benchmarks. For the first task, they masked only entity names with a special token (\textit{masked-entity}; ME), whereas for the second task, only the entity names were given (\textit{only-entity}; OE). The model maintained higher performances in the OE setting, indicating that the entity names were more informative of the predicted relation than the contextual information. This hints at what the authors call \textit{semantic bias}.

\paragraph{A Note on Reporting Bias}
Generally, when extracting knowledge from text, one should be aware that the frequency with which facts are reported is not representative of their real-world prevalence. Humans tend to mention only events, outcomes, or properties that are out of their perceived ordinary \citep{gordon_reporting_2013} (e.g., "a banana is yellow" is too trivial to be reported). This phenomenon is called \textit{reporting bias} and likely stems from a need to be as informative and non-redundant as possible when sharing knowledge. 

\section{Bias in Knowledge Graphs}
\label{bias_in_KG}

Next in our investigation of the lifecycle (Figure \ref{fig:schema_lifecycle}) comes the representation of entities and relations as a KG. In the following, we illustrate which social biases are manifested in KGs and how.  

\subsection{Descriptive Statistics}
\citet{janowicz_debiasing_2018} demonstrated that DBpedia, which is sourced from Wikipedia info boxes, mostly represents the western and industrialized world. Matching the coverage of location entries in the KG with population density all over the world showed that several countries and continents are underrepresented. 
A disproportionate 70\% of the person entities in Wikidata are male (20\% are female, less than 1\% are neither male nor female, and for roughly 10\% the gender is not indicated) \citep{beytia2021visual}. 
 \citet{RadstokCS21} found that the most frequent occupation is \textit{researcher} and \citet{beytia2021visual} identified \textit{arts}, \textit{sports}, and \textit{science and technology} as the most prominent occupation categories. In reality, only about 2\% of people in the U.S. are researchers \citep{RadstokCS21}. This gap is likely caused by reporting bias as discussed earlier (Section \ref{auto_info_extraction}). \citet{RadstokCS21}, moreover, observed that mentions of ethnic group membership decreased and changed in focus between the 18th and 21st century. Greeks are the most frequently labeled ethnic group among historic entries (over 400 times) and African Americans among modern entries (only roughly 100 times).

\subsection{Semantic Polarity}
\citet{mehrabi_lawyers_2021} focused on biases in common sense KGs like ConceptNet \citep{DBLP:conf/aaai/SpeerCH17} and GenericsKB \citep{genericskb} (contains sentences) which are at risk of causing representational harms (see Section \ref{harms}). They utilized \textit{regard} \citep{sheng-etal-2019-woman} and \textit{sentiment} as intermediate bias proxies. Both concepts express the polarity of statements and can be measured via classifiers that predict a neutral, negative, or positive label  \citep{sheng-etal-2019-woman, dhamala-2021-bold}. Groups that are referred to in a mostly positive way are interpreted as favored and vice versa. \citet{mehrabi_lawyers_2021} applied this principle to natural language statements generated from ConceptNet triples.  
They found that subject and object entities relating to the professions \textit{CEO}, \textit{nurse}, and \textit{physician} were more often favored while \textit{performing artist}, \textit{politician}, and \textit{prisoner} were more often disfavored. Similarly, several Islam-related entities were on the negative end while \textit{Christian} and \textit{Hindu} were more ambiguously valuated. As for gender, no significant difference was found.

\section{Bias in Knowledge Graph Embeddings}
\label{bias-in-KGE}
 Vector representations of KGs are used in a range of downstream tasks or combined with other types of neural models \citep{DBLP:journals/pieee/Nickel0TG16,DBLP:journals/semweb/RistoskiRNLP19}. They facilitate efficient aggregation of connectivity patterns and convey latent information.

Embeddings are created through statistical modeling and summarize distributional characteristics. So, if a KG like Wikidata contains mostly (if not only) male presidents, the relationship between the gender \textit{male} and the profession \textit{president} is assumed to manifest itself accordingly in the model. In fact, the papers summarized below provide evidence that the social biases of KGs are modeled or further amplified by KG embeddings (KGEs).  The following sections are organized by measurement strategy to give an overview of existing approaches and the information gained from them.

\begin{table*}[t]
\scriptsize
\centering
\caption{Overview of reviewed works concerning the sources, measurement, and mitigation of bias in KGs/KGEs.}
\begin{tabular}{lll}
\label{emb-bias-overview}
\textbf{Bias Source}   & & \\\hline
    Crowd-Sourcing & & \citet{beytia2021visual,janowicz_debiasing_2018,demartini_implicit_2019} \\
    Ontologies & & \citet{janowicz_debiasing_2018,keet2021exploration,paparidis-towards-2021,geller-2021-detecting} \\
    Extraction & &\citet{mehrabi_man_2020,mishra_assessing_2020,gaut_towards_2020,li2021robustness} \\\hline 
\textbf{Bias Measurement}   &  & \\\hline
    \textit{Representation} & \textit{Method} &
    \\
    KG & Descriptive Statistics    &       \citet{janowicz_debiasing_2018,RadstokCS21,beytia2021visual}   \\
    & Semantic Polarity                    &       \citet{mehrabi_lawyers_2021}    \\\hline
    KGE              & Analogies                   &       \citet{bourli_bias_2020}  \\
    & Projection                  &       \citet{bourli_bias_2020}  \\
    & Update-Based   &       \citet{fisher_measuring_2020,keidar_towards_2021,du-etal-2022-understanding} \\
    & Link Prediction             &   \citet{keidar_towards_2021,arduini_adversarial_2021,RadstokCS21,du-etal-2022-understanding} \\ \hline
\textbf{Bias Mitigation}    &   &  \\\hline
\textit{Representation} & \textit{Method} & \\
    KGE & Data Balancing           &   \citet{RadstokCS21,du-etal-2022-understanding} \\
    & Adversarial Learning        &   \citet{fisher_debiasing_2020,arduini_adversarial_2021} \\
    & Hard Debiasing              &   \citet{bourli_bias_2020}  
\end{tabular}
\end{table*}

\subsection{Stereotypical Analogies}
The idea behind analogy tests is to see whether demographics are associated with attributes in stereotypical ways (e.g., "Man is to computer programmer as woman is to homemaker" \citep{bolukbasi2016man}). In their in-depth analysis of a TransE-embedded Wikidata KG, \citet{bourli_bias_2020} investigated occupational analogies for binary gender seeds. TransE \citep{bordes-2013-transe} represents $(h,r,t)$ (with head $h$, relation $r$, tail $t$) in a single space such that $h+r\approx t$. The authors identified the model's most likely instance of the claim "\textit{a} is to \textit{x} as \textit{b} is to \textit{y}" (with \textit{(a,b)} being a set of demographics seeds and \textit{(x,y)} a set of attributes) via a cosine score: $S_{(a,b)}(x,y)=cos(\vec{a}+\vec{r}-\vec{b},\vec{x}+\vec{r}-\vec{y})$, where $r$ is the relation \textit{has\_occupation}.
In their study, the highest scoring analogy was "woman is to fashion model as man is to businessperson". This example appears rather stereotypical, but other highly ranked analogies less so, like "Japanese entertainer" versus "businessperson" \citep{bourli_bias_2020}. A systematic evaluation of how stereotypical the results are is missing here. In comparison, the work that originally introduced analogy testing for word2vec \citep{bolukbasi2016man} employed human annotators to rate stereotypical and gender-appropriate analogies (e.g., "sister" versus "brother"). 

\subsection{Projection onto a Bias Subspace}
\label{projection-measure}
Projection-based measurement of bias is another approach that was first proposed by \citet{bolukbasi2016man} for word embeddings, and was adapted for TransE by \citet{bourli_bias_2020}. In a first step, a one-dimensional gender direction $\vec{d}_g$ is extracted. Then, a projection score metric $S$ is computed to indicate gender bias --- with projection $\pi$ of an occupation vector $\vec{o}$ onto $\vec{d}_g$ and a set of occupations $C$: $S(C)=\frac{1}{|C|}\sum_{o\in C}||\pi_{\vec{d}_g} \vec{o}||$. Occupations with higher scores are interpreted as more gender-biased and those with close-to-zero scores as neutral.

\subsection{Update-Based Measurement}
\label{tl-measure}
The \textit{translational likelihood} (TL) metric was tailored for translation-based modeling approaches \citep{fisher_measuring_2020}. To compute this metric, the embedding of a person entity is updated for one step towards one pole of a seed dimension. This update is done in the same way as the model was originally fit in. For example, if head entity \textit{person x} is updated in the direction of \textit{male} gender, the TL value is given by the difference between the likelihood of \textit{person x} being a \textit{doctor} after versus before the update. If the absolute value averaged across all human entities is high, this indicates a bias regarding the examined seed-attribute pair.
\citet{fisher_measuring_2020} argue that this measurement technique avoids model-specificity as it generalizes to any scoring function. However, \citet{keidar_towards_2021} found that the TL metric does not compare well between different types of embeddings (details in Section \ref{link-pred-measure}). It should, thus, only be used for the comparison of biases within one kind of representation. 
\citet{du-etal-2022-understanding} propose an approach comparable to \citet{fisher_measuring_2020} to measure individual-level bias. Instead of updating towards a gender dimension, the authors suggest flipping the entity's gender and fully re-training the model afterward. The difference between pre- and post-update link prediction errors gives the bias metric. A validation of the approach was done on TransE for a Freebase subset (FB5M \citep{DBLP:journals/corr/BordesUCW15}) \citep{du-etal-2022-understanding}. The summed per-gender averages (group-level metric) were found to correlate with U.S. census gender distributions of occupations. 

\section{Downstream Task Bias: Link Prediction}
\label{link-pred-measure}
Link prediction is a standard downstream task that targets the prediction of relations between entities in a given KG. Systematic deviations in the relations suggested for entities with different demographics indicate reproduced social bias.

For the measurement of fairness or bias in link prediction, \citet{keidar_towards_2021} distinguish between \textit{demographic parity} versus \textit{predictive parity}. The assumption underlying demographic parity is that the equality between predictions for demographic counterfactuals (opposite demographics, for example, \textit{female} versus \textit{male} in binary understanding) is the ideal state \citep{dwork2012fairness}. That is, the probability of predicting a label should be the same for both groups.  
Predictive parity is given, on the other hand, if the probability of true positive predictions (\textit{positive predictive value} or \textit{precision}) is equal between groups \citep{chouldechova2017fair}. Hence, this measure factors in the label distribution by demographic. 
With these metrics, \citet{keidar_towards_2021} analyzed different embedding types, namely TransE, ComplEx, RotatE, and DistMult, each fit on the benchmark datasets FB15k-237 \citep{toutanova-chen-2015-observed} and Wikidata5m \citep{wang2021kepler}. They averaged the scores across a large set of human-associated relations to detect automatically which relations are most biased. The results showed that \textit{position played on a sports team} was most consistently gender-biased across embeddings. \citet{arduini_adversarial_2021} analyzed link prediction parity regarding the relations \textit{gender} and \textit{occupation} to estimate debiasing effects on TransH \citep{Wang_Zhang_Feng_Chen_2014} and TransD \citep{ji-etal-2015-knowledge}. The comparability between different forms of vector representations is a strength of downstream metrics. In contrast, measures like the analogy test or projection score \citep{bourli_bias_2020} are based on specific distance metrics and TL \citep{fisher_measuring_2020} was shown to lack transferability across representations \citep{keidar_towards_2021} (Section \ref{tl-measure}).

\citet{du-etal-2022-understanding} interpret the correlation between gender and link prediction errors as an indicator of group bias. With this, they found, for example, that \textit{engineer} and \textit{nurse} are stereotypically biased in FB5M. However, the ground truth gender ratio was found not predictive of the bias metric (e.g., despite its higher male ratio, \textit{animator} produced a stronger female bias value). For validation, it was shown that the predicted bias values correlate to the gender distributions of occupations according to U.S. census (again, on TransE). Furthermore, the authors investigated how much single triples contribute to group bias via an \textit{influence function}. They found that gender bias is mostly driven by triples containing gendered entities and triples of low degree.

\section{Breaking the Cycle? Bias Mitigation in Knowledge Graph Embeddings}
\label{bias-mitigation-in-KGE}
A number of works have attempted to post-hoc mitigate biases in KGEs. Given that pre-existing biases are hard to eradicate from KGs, manipulating embedding procedures, may alleviate the issue at least on a representation level. In the following, we summarize respective approaches.

\subsection{Data Balancing}
\label{balanced-sampling}

\citet{RadstokCS21} explored the effects of training an embedding model on a gender-balanced subset of Wikidata triples. First, the authors worked with the originally gender-imbalanced Wikidata12k \citep{DBLP:conf/www/LeblayC18,DBLP:conf/emnlp/DasguptaRT18} and DBpedia15k \citep{DBLP:conf/semweb/SunHL17} on which they fit a TransE and a DistMult model \citep{distmult}. They then added more female triples from the Wikidata/DBpedia graph to even out the binary gender distribution among the top-5 most common occupations. Through link prediction, they compared the number of male and female predictions with the ground truth frequencies. More female entities were predicted after the data balancing intervention. However, the absolute difference between the female ratios in the data and the predictions increased, causing the model to be less accurate and fair. Moreover, the authors note that this process is not scalable since for some domains there are no or only a limited amount of female entities (e.g., female U.S. presidents do not exist in Wikidata). 

\citet{du-etal-2022-understanding} experimented with adding and removing triples to gender-balance a Freebase subset \citep{DBLP:journals/corr/BordesUCW15}. For the first approach, the authors added synthetic triples (as opposed to real entities from another source as was done by \citet{RadstokCS21}) for occupations with a higher male ratio. The resulting bias change was inconsistent across occupations. This appears in line with the authors' finding that ground truth gender ratios are not perfectly predictive of downstream task bias (Section \ref{link-pred-measure}). For the second strategy, the triples that most strongly influenced an existing bias were determined and removed. This outperformed random triple removal.

\subsection{Adversarial Learning}
\label{adv_learning}
 Adversarial learning for model fairness aims to prevent prediction of a specific personal attribute from a person's entity embedding. As an adversarial loss, \citet{fisher_debiasing_2020} used the KL-divergence between the link prediction score distribution and an idealized target distribution. For example, for an even target score distribution for a set of religions, the model is incentivized to give each of them equal probability. However, in their experiments, this treatment failed to remove the targeted bias fully. This is likely caused by related information encoded in the embedding that is able to inform the same bias.
 
 \citet{arduini_adversarial_2021} used a Filtering Adversarial Network (FAN) with a filter and a discriminator module. The filter intends to remove sensitive attribute information from the input, while the discriminator tries to predict the sensitive attribute from the output. Both modules were separately pre-trained (filter as an identity mapper of the embedding and discriminator as a gender predictor) and then jointly trained as adversaries. In their experiments, the gender classification accuracy for high- and low-degree entities was close to random for the filtered embeddings (TransH and TransD). For an additional occupation classifier, accuracy remained unaffected after treatment.
 
 \subsection{Hard Debiasing}
 \label{hard_debias}
 \citet{bourli_bias_2020} propose applying the projection-based approach explained in Section \ref{projection-measure} for the debiasing of TransE occupation embeddings. To achieve this, its linear projection onto the previously computed gender direction is subtracted from the occupation embedding. A variant of this technique ("soft" debiasing) aims to preserve some degree of gender information by applying a weight $0 < \lambda < 1$  to the projection value before subtraction. In the authors' experiments, the correlation between gender and occupation was effectively removed --- as indicated by the projection measure \citep{bourli_bias_2020}. However, the debiasing degree determined by $\lambda$ was found to be in trade-off with model accuracy. 
 This technique was closely adapted from \citet{bolukbasi2016man}, regarding which \citet{gonen2019lipstick} criticize that gender bias is only reduced according to their specific measure and not the "complete manifestation of this bias".

\section{Discussion}

 In this article, we cover a wide range of evidence for harmful biases at different stages during the lifecycle of "facts" as represented in KGs. Some of the most influential graphs misrepresent \textit{the world as it is} due to sampling and algorithmic biases at the creation step. Pre-existing biases are exaggerated in these representations. Embedding models learn to encode the same or further amplified versions of these biases. Since the training of high-quality embeddings is costly, they are, in practice, pre-trained once and afterward reused and fine-tuned for different systems. These systems preserve the inherited biases over long periods, exacerbating the issue further. Our survey shows that KGs may qualify as resources for historic facts, but they do not qualify for inference regarding various human attributes. Future work on biases in KGs and KGEs should aim for improvement in the following areas:

\paragraph{Attribute and Seed Choices} Bias metrics usually examine one or a few specific attributes (e.g., occupation) and their correlations with selected seed dimensions (e.g., gender). Occupation is by far the most researched attribute in the articles we found \citep{arduini_adversarial_2021,RadstokCS21,bourli_bias_2020, fisher_debiasing_2020, fisher_measuring_2020}. Only \citet{keidar_towards_2021} propose to aggregate the correlations between a set of seed dimensions and all relations in a graph. All the works used binary gender as the seed dimension and some additionally addressed ethnicity, religion, and nationality \citep{fisher_debiasing_2020,fisher_measuring_2020,mehrabi_lawyers_2021}. 

\paragraph{Lack of Validation}
Most of the KGE bias metrics presented here are interpreted as valid if they detect unfairly discriminating association patterns that intuitively align with existing stereotypes. Besides that, several works investigate the comparability between different metrics. Although both of these practices deliver valuable information on validity, they largely ignore the societal context. Only \citet{du-etal-2022-understanding} compared embedding-level bias metrics with census-aligned data to assess compatibility with real-world inequalities. We suggest that future work consider a more comprehensive study of \textit{construct validity} (Does the measurement instrument measure the construct in a meaningful and useful capacity?) \citep{jacobs-2021-measurement}. One requirement is that the obtained measurements capture all relevant aspects of the construct the instrument claims to measure. That is, a gender bias measure must measure all relevant aspects of gender bias \citep{stanczak2021survey} (including, e.g., nonbinary gender and a distinction between benevolent and hostile forms of sexist stereotyping \citep{glick1997hostile}). Unless proven otherwise, we must be skeptical that this is achieved by existing approaches \citep{gonen2019lipstick}.
As a result of minimal validation, detailed interpretation guidelines are generally not provided. Therefore, the distinctions between strong and weak bias or weak bias and random variation are mostly vague. 

\paragraph{(In-)Effectiveness of Mitigation Strategies}

Data balancing is the most intuitive approach to bias mitigation and was proven to be effective in the context of text processing \citep{DBLP:conf/acl/MeadePR22}. However, for KGEs, data balancing methods were found to inconsistently reduce bias (Section \ref{balanced-sampling}). Adversarial learning yielded promising outcomes in the study by \citet{arduini_adversarial_2021}. Their FAN approach does not rely on pre-specified attributes. This is in contrast to \citet{fisher_debiasing_2020}, whose intervention was found to miss non-targeted, yet bias-related information. This problem relates to one of the main criticisms of hard and soft debiasing: instead of alleviating the problem, these techniques risk concealing the full extent of the bias \citep{gonen2019lipstick}.

\paragraph{Reported Motivations}
Many, yet not all works in the field name potential social harms as a motivator for their research on social bias in KGs \citep{mehrabi_lawyers_2021,fisher_debiasing_2020, fisher_measuring_2020,RadstokCS21}. Only \citet{mehrabi_lawyers_2021} drew from established taxonomies and targeted biases associated with \textit{representational harms} \citep[Barocas et al., as cited in,][]{blodgett2020language}. Similarly, most works lack a clear working definition of social bias. For example, aspects of pre-existing societal biases captured in the data and biases arising through the algorithm \citep{friedmann1996bias} are usually not disentangled. Only \citet{bourli_bias_2020} compared model bias to the original KG frequencies and showed that the statistical modeling caused an amplification. 

\section{Recommendations}
To avoid harms caused by biases in KGs and their embeddings, we identify and recommend several actions for practitioners and researchers. 

\paragraph{Transparency and Accountability}
KGs should by default be published with bias-sensitive documentation to facilitate transparency and accountability regarding potential risks. \textit{Data Statements} \citep{bender-friedman-2018-data} report  curation criteria, language variety, demographics of the data authors and annotators, relevant indicators of context, quality, and provenance. \textit{Datasheets for Datasets} \citep{DBLP:journals/cacm/GebruMVVWDC21} 
additionally state motivation, composition, preparation, distribution, and maintenance. The associated questionnaire can accompany the dataset creation process to avoid risks early on. Especially in the case of ongoing crowd-sourcing efforts for encyclopedic KGs the demographic background of  contributors should be reported \citep{demartini_implicit_2019}. Researchers using subsets of these KGs, should investigate respective data dumps for potential biases and report limitations transparently.  Similarly, KG embedding models should be published with \textit{Model Cards} \citep{mitchell2019model} documenting intended use, underlying data, ethical considerations, and limitations. 
Stating the contact details for reporting problems and concerns establishes accountability \citep{mitchell2019model,DBLP:journals/cacm/GebruMVVWDC21}.

\paragraph{Improving Representativeness}
To tackle selection bias, data collection should aim to employ authors and annotators from diverse social groups and with varied cultural imprints. Annotations should be determined via aggregation \citep[see][]{DBLP:journals/llc/HovyP21}. For open editable KGs, 
interventions like \textit{edit-a-thons} are helpful to introduce more authors from underrepresented groups \citep{vetter2022assessing} (e.g., the Art+Feminism campaign  aims to fill the gender gap in Wikimedia knowledge bases\footnote{\url{https://outreachdashboard.wmflabs.org/campaigns/artfeminism_2022/overview}}). In order for such interventions to take effect, research must update data bases and benchmarks frequently \citep[see][]{DBLP:conf/nips/KochDHF21}. In addition, the timeliness of encyclopedic data is necessary to avoid perpetuating historic biases. 

\paragraph{Tackling Algorithmic Bias}
Evaluation and prevention of harmful biases must become part of the development pipeline \cite{stanczak2021survey}. 
 Algorithmic biases are best evaluated with a combination of multiple quantitative (Section \ref{bias-in-KGE})  and qualitative measures \citep[][]{mci/Kraft2022,Dev2021What}, considering multiple demographic dimensions (beyond gender and occupation). Evaluating the content of attributions in light of social discourse and the intended use of a technology facilitates an assessment of potential harms \citep{selbst2019fairness}. Downstream task bias may exist independently from a measured embedding bias \citep{goldfarb-tarrant-etal-2021-intrinsic}, therefore a task- and context-oriented evaluation is preferred (Section \ref{link-pred-measure}).
 We have presented several bias-mitigating strategies for different KGEs, which might alleviate the issue in some cases (Section \ref{bias-mitigation-in-KGE}). However, more research is needed to establish more effective and robust mitigation methods, as well as metrics used to evaluate their impact \citep{gonen2019lipstick,blodgett2020language}.

\section{Related Work}
Although a wide range of surveys investigates biases in NLP, none of them addresses KG-based methods, in particular. \citet{blodgett2020language} critically investigated the theoretical foundation of works analyzing bias in NLP. The authors claim that most works lack a clear taxonomy. We came to a similar conclusion with respect to evaluations of KGs and their embeddings. \citet{sun-etal-2019-mitigating} and \citet{stanczak2021survey} surveyed algorithmic measurement and mitigation strategies for gender bias in NLP. \citet{sheng-etal-2021-societal} summarized approaches for the measurement and mitigation of bias in generative language models. Some of the methods presented earlier are derived from works discussed in these surveys and adapted to the constraints of KG embeddings (e.g., \citet{bourli_bias_2020} adapted hard debiasing \citep{bolukbasi2016man}). Criticisms point to the monolingual focus on the English language, the predominant assumption of a gender binary, and a lack of interdisciplinary collaboration.

 \citet{shah-etal-2020-predictive} identified four sources of predictive biases: \textit{label bias} (label distributions are imbalanced and erroneous regarding certain demographics), \textit{selection bias} (the data sample is not representative of the real world distribution), \textit{semantic bias/input representation bias} (e.g., feature creation with biased embeddings), and \textit{overamplification} through the predictive model (slight differences between human attributes are overemphasized by the model). All of these factors are reflected in the lifecycle as discussed in this article. To counter the risks,  \citet{shah-etal-2020-predictive}  suggest employing multiple annotators and methods of aggregation \citep[see also][]{DBLP:journals/llc/HovyP21}, re-stratification, re-weighting, or data augmentation, debiasing of models, and, finally, standardized data and model documentation.

\section{Conclusion and Paths Forward}
Our survey shows that biases affect KGs at different stages of their lifecycle. Social biases enter KGs in various ways at the creation step (e.g., through crowd-sourcing of triples and ontologies) and manifest in popular graphs, like DBpedia \citep{beytia2021visual} or ConceptNet \citep{mehrabi_lawyers_2021}. Embedding models can capture exaggerated versions of these biases \citep{bourli_bias_2020}, which finally propagate downstream  \citep{keidar_towards_2021}. 
We acknowledge that KGs have enormous potential for a variety of knowledge-driven downstream applications \cite{DBLP:journals/semweb/Martinez-Rodriguez20, DBLP:journals/ki/NgomoSGHDLOS21,DBLP:conf/sigir/JiangU22} and improvements in the truthfulness of statistical models \cite{DBLP:conf/www/AthreyaNU18,DBLP:journals/corr/abs-2204-09149}. Yet, although KGs are factual about historic instances, they also perpetuate historically emerging social inequalities.
Thus, ethical implications must be considered when developing or reusing these technologies.

We showed that most embedding-based measurement approaches for bias are still restricted to a limited number of demographic seeds and attributes. Furthermore, their alignment with social bias as a construct is not sufficiently validated. Some debiasing strategies appear effective within rather narrow definitions of bias. More in-depth scrutiny is required for a broader understanding of bias. Future work should be grounded in an  investigation of concepts like gender or ethnic bias and strive for more comprehensive operationalizations and validation studies. Finally, the motivations and conceptualizations should be communicated clearly.

\section*{Acknowledgments}
We acknowledge the financial support from the Federal Ministry for Economic Affairs and Energy of Germany in the project CoyPu (project number 01MK21007[G]) and the German Research Foundation in the project NFDI4DS (project number 460234259).

\bibliography{custom}

\appendix

\end{document}